1

# Evolution Theory of Self-Evolving Autonomous Problem Solving Systems

Seppo Ilari Tirri

*Information & Communication Technology*
*Asia e University (AeU), Malaysia*
*Jalan Sultan Sulaiman, 50000 Kuala Lumpur, Malaysia*

`seppo.tirri@aeu.edu.my`

ABSTRACT

The present study gives a mathematical framework for self-evolution within autonomous problem solving systems. Special attention is set on universal abstraction, thereof generation by net block homomorphism, consequently multiple order solving systems and the overall decidability of the set of the solutions. By overlapping presentation of nets new abstraction relation among nets is formulated alongside with consequent alphabetical net block renetting system proportional to normal forms of renetting systems regarding the operational power. A new structure in self-evolving problem solving is established via saturation by groups of equivalence relations and iterative closures of generated quotient transducer algebras over the whole evolution.

# INTRODUCTION

The present study is a continuation work of my previous work in the art:" Algebraic Net Class Rewriting Systems, Syntax and Semantics for Knowledge Representation and Automated Problem Solving" in Tirri SI (2013), and preliminaries as well as related notations are to be found there.

Lots of studies have been driven to clarify routes between nodes e.g. in process algebra, important topics setting ground to game theories as well as overall in halting problems. On the other hand in more complex dimensional cases ordering definitions in sets of subgraphs have been under vigorous investigations mainly concentrated in tree structures. An amazingly minute portion of studies on graphs concentrates to relations between graphs and abstraction of them and one explanation for this might be that transformations on conceptual levels lead joints to a succinct model proper to syntax as well as to semantic domain requiring combining algebraic structures to loop structured graphs and realizations of them, this requiring symbiosis of abstract syntax and real case sides.

The most remarkable study of human abstraction mechanism yielding a concrete result especially within mathematics in the form of analytical tools has been manifested by French philosopher, mathematician and physicist René Descartes in the 17th century in his work "*Regulae ad directionem ingenii*, *Règles utiles et claires pour la direction de l'Esprit en la recherche de la Vérité* (1628)", freely outlining: "… at first we must organize the things which are the most essential ones in concentrating to do that by simplifying from phase to phase the vague, indefinite original problem. Then we try to understand the relations between those simplified parts and then compare the propositions to be proved i.e. wise versa try to see the connections between the reached relations and the original problem....". Descartes underlines the importance of the origin of deduction itself namely abstraction by stating "there is not very much in results or even in the proofs of them, but the method how they are invented, that is what is the process inventors use to realize proofs". The first beneficial syntax for the use of infinite sequences in calculus was realized by Gottfried Leibniz in his infinitesimal approaches applied in analysis in geometry, but formalizing reasoning accelerated not until the breakthrough ideas of Alan Turing on string language representation methods in 1930s, which adopted in practice during 40s and 50s when the main target was to speed up data handling mainly in the near branch. However this was to push aside the previously nascent contemplation about the essence of reasoning itself and mathematically modeling notion abstraction was to be postponed. After technological revolution having achieved sufficiently strength in 80s Japan has been performing as the most driving force to implement applications which would be prominent vehicles for executing more extensive mathematical reasoning in a variety of situations by constructing robots. On uppermost have been expert systems and imitation of human actions example wise without innovations towards inventiveness processes behind novel reasoning. Prospective preliminary approaches by implicit abstraction formulation are raised by Plaisted (1981) "clauses mapping" between ground theories and abstract counterparts i.e. relations between syntactical predicates and their evaluations and their generalizations to operator evaluations Nayak PP, Levy AY (1995) and "ground language" corresponding to a Boolean graph over variables and a set of ground formulas over possibly more extensive set of variables Giordana A, Saitta L (1990) and abstraction between the whole formal structures: languages, axioms and inference rules Giunchiglia F, Walsh T (1992) and "grounded abstraction" combining the whole variety of conceptualized ground entities Saitta L, Zucker J-M (2001). However the implicit nature of those existing models are not sufficiently expressive considering applications on knowledge representation and exact algebraic formalization combining abstraction operators and their semantic counterparts has been waiting for emerging.

The question in automated problem solving basically is how to generate nets from enclosements of a probed net those enclosements being in such a relation with the enclosements in the conceptual nets that the particular relation is invariant under that generating transformation i.e. preserves invariability under class-rewriting. Each perception as itself is able to orchestrate only a rough depict of problem subject under investigation but as posing a conceptual representation and parallelled with other already known concepts liable to the same subject is able to offer an explicit gateway for self-evolutional solving systems via algebraic quotient rewriting closure methods. Therefore we handle an idea of automated problem solving as formal inventiveness. In problem solving an essential thing is to see over details, and that is the task we next grip ourselves into by describing ideas such as partitioning nets by normal forms of renetting systems and a connection between partitions by introducing the abstraction relation. We concentrate to construct TD-models for formulas of jungle pairs by conceptualizing ground subjects and then reversing counterparts of existing TD-solutions back to ground level. Then we widen the solution hunting by taking universal partitioning into action in order to allow new links to the environment of applicants. By widening net homomorphism to cover idea of block-altering we obtain TD-transformation generation sets and as a consequence coherent expansion to next order solution levels. Saturation by groups of equivalence relations consistently supplements TD-solution arsenal and by iterating alternately mother nets and solutions in solving system leads to multidimensional infinite solving process. Furthermore new solutions will be added to the set of already known ones thus expanding the solving power in the forthcoming - hereby establishing a self-evolving autonomous learning system.

The present study use notion *net* and its operational counterpart for a basic explicit mathematical formalism for syntax of generic notion of knowledge and its semantics for real world cases by offering an algebraic approach to implement consecutive simultaneously looped deterministic and undeterministic operations in generalized universal free algebra and its realizations on different algebras, cf. traditional handling for trees and the evaluations Burris S, Sankappanavar HP (1981), Ohlebusch E (2002) and Denecke K, Wismat SL (2002). Nets can be identified unequivocally by any member of the corresponding net class by root basis the most appropriate to the occasions. In syntax point of view "hyperedges" Engelfriet J (1997) can be regarded as nodes with in- and outarities of nets and in semantic aspect edge graphs equate with realization process graphs with nodes related to in-/outputs and edges to transformation relations.

The graph transformations from operator to operator are extended to cover more complex cases than appear in trees and to be more proper as premises of substitutions considering matching properties in general rewriting systems. *Net substitutions* as a special case of more generic net homomorphisms are used to nicely manifest a unified way to express replacements in complicated iterative network in *universal partitioning* compared to the more limited version used in trees Jantzen M (1997).

Graph rewriting are mostly seen to be defined with a sample of rules between elementary entities as edges and nodes thus serving only as implicit tools, not as explicit ones needed for handling more extensive graph structures as a whole. Therefore *Renetting systems* are established as rewriting systems constructed to be adaptable to succinct algebraically represented graphs as well as trees thus avoiding weaker implicit expressions by rules based on exclusively sets of edges and vertices. Renetting systems are equipped with specifics incorporating accounts of the positions in targeted subnets and also the differences between left and right substitutions as well as other more traditional limit demands relating to applied rule orders or simultaneity inter alia cf. Ohlebusch E (2002), Engelfriet J (1997), "priority" by Baeten JCM, Basten T (2001) and Cleaveland R et al. (2001) "probability" Jonsson B et al. (2001). By intervening renetting systems being orchestrated to deploy new binding organizations to environments of applied entities we expand notional perception to cases where the common origins with already known conceptual counterparts are allowed to possess more limited interface consequently increasing the possibility to find suitable transformation rule sets as solution candidates for target perception entities.

*Transducers* are net realizations with renetting systems as operation vertexes and serve as groups pair wise commuting parallel operations on abstract class quotient algebra composing the closures of solving algorithm structures.

The present study serves as an explicit algebraic system for generic knowledge constitution and problem solving closure structures derived from abstraction classes and class renetting systems over them. Nets are ideal constructions maintaining prime features of operators in intervening rewriting corresponding under special issues of abstract classes manifested by said rewriting to "grounded abstraction" Saitta L, Zucker J-D (2001), where interpretations are organized by shifts inside partition net classes possessing a common origin and in single cases from algebra to its free algebra syntax; consequently widening grounded abstraction notion to deal net class abstraction instead of only single perception cases. Parallel algebraic transducers serve as more general and explicit counterparts to solution memorizing operators as well as intervening rewriting systems stand for abstract perception operators.

In problem solving area parallel class rewriting covers block graph transformations and the idea of partially matching while net rewriting itself is a generalization to "genetic algorithms" Negnevitsky M (2002) which can be explained as term rewriting within horizontal changes on leaves (terminal letters), in contents as well as in arrangements; nets themselves offering in the realization aspect up-coming streams to influence to the results of operator realizations.

METHOD AND TARGET

By changing indexing inside nets net block is presented to expand net handling alternatives regarding reorganization into environments. *Net block homomorphism* can be determined to serve as initial abstraction operation via replacing the sets of blocks by letters. By alphabetical net block homomorphism new abstraction relation over the set of the nets is established. As the ground of autonomous solving evolution multiple level abstract algebra equivalent class transducer as operations are constructed of which consequently iteration. From the basis of alphabetical net block homomorphism new renetting systems are obtained to generate normal forms of any renetting system. Saturation by groups of equivalence classes offers next abstraction level to deal targets in the light of wider comprehension. Introducing *multilevel iterative abstract quotient transducer classes* and assuming mother nets and known solutions be fixed in each level and extending nested processes further exponentially we´ll get autonomous evolution levels and obtain transducer equivalent classes alteration to upgraded abstraction levels and consequently applying the whole string of the achieved process finally manifests self-evolving unrestricted autonomous problem solving formalism.

1§                    Preliminaries

First we recall some keen definitions.

*Net Transformations*

NET HOMOMORPHISM. Let X and Y be frontier alphabets, $\Sigma$ and $\Omega$ ranked alphabets and $\Xi_\Sigma$ and $\Xi_\Omega$ arity alphabets, especially distinct rank-indexed arity alphabets $E_{in} = \{\varepsilon_i : i \in \mathcal{E}_{in}\}$ for in-arities and $E_{out} = \{\varepsilon_i : i \in \mathcal{E}_{out}\}$ for out-arities respectively, disjoint from all other used alphabets.

*Net homomorphism* h: $F_\Sigma(X,\Xi_\Sigma) \cup \widetilde{F}_\Sigma(X,\Xi_\Sigma) \mapsto P(F_\Omega(Y,\Xi_\Omega) \cup \widetilde{F}_\Omega(Y,\Xi_\Omega))$ is a relation where

$h(t) = h_\Sigma(\sigma)(h(\mu_i); h(\lambda_j) \mid i \in \mathcal{E}_{inh_\Sigma(\sigma)}, j \in \mathcal{E}_{outh_\Sigma(\sigma)})$ , whenever $t = \sigma(\mu_i; \lambda_j \mid i \in \mathcal{I}_\sigma, j \in \mathcal{J}_\sigma) \in F_\Sigma(X,\Xi_\Sigma)$,

and h: $\Sigma_0 \cup X \cup \Xi_\Sigma \mapsto P(\widetilde{F}_\Omega(Y,\Xi_\Omega)^{(1)} \cup \Omega_0 \cup \Xi_\Omega)$, where

for each $\xi \in \Xi_\Sigma$ $h(\xi) \in P(\Xi_\Omega)$ and $h(\sigma) \in \Omega_0$ whenever $\sigma \in \Sigma_0$ ;

h|X named the *initial manoeuvre rewriting relation*, and h|$\Xi_\Sigma$ the *initial arity rewriting relation* ;

$h_\Sigma : \Sigma \mapsto P(F_\Omega(Y, \Xi_\Omega \cup E_{in} \cup E_{out}) \cup \Omega)$ is a $\Sigma$-*ranked letter rewriting relation* ,

$h(u) \in P(\widetilde{F}_{in\Omega}(Y,\Xi_\Omega)^{(1)})$, and $h(u)_L = h(u_L)$, whenever $u \in \widetilde{F}_{in\Sigma}(X, \Xi_\Sigma)^{(1)}$ and $u_L \in \Sigma_0$,

$h(u) \in P(\widetilde{F}_{out\Omega}(Y,\Xi_\Omega)^{(1)})$, and $h(u)_L = h(u_L)$, whenever $u \in \widetilde{F}_{out\Sigma}(X,\Xi_\Sigma)^{(1)}$ and $u_L \in \Sigma_0$,

$h(u) \in P(\widetilde{F}_{in\Omega}(Y,\Xi_\Omega)^{(1)} \cup \widetilde{F}_{in\Omega}(Y,\Xi_\Omega)^{(2)})$, whenever $u \in \widetilde{F}_{in\Sigma}(X,\Xi_\Sigma)^{(1)}$ and $u_L \in X$,

$h(u) \in P(\widetilde{F}_{out\Omega}(Y,\Xi_\Omega)^{(1)} \cup \widetilde{F}_{out\Omega}(Y,\Xi_\Omega)^{(2)})$, whenever $u \in \widetilde{F}_{out\Sigma}(X,\Xi_\Sigma)^{(1)}$ and $u_L \in X$,

$h(u) \in P(\widetilde{F}_{in\Omega}(Y,\Xi_\Omega)^{(2)})$, and $h(u)_L = h(u_L)$, whenever $u \in \widetilde{F}_{in\Sigma}(X,\Xi_\Sigma)^{(2)}$, and

$h(u) \in P(\widetilde{F}_{out\Omega}(Y,\Xi_\Omega)^{(2)})$, and $h(u)_L = h(u_L)$, whenever $u \in \widetilde{F}_{out\Sigma}(X,\Xi_\Sigma)^{(2)}$.

*Net substitution relation* (here f) is such a special case of net homomorphism in $F_\Sigma(X,\Xi) \cup \widetilde{F}_\Sigma(X,\Xi)$ that each ranked letter rewriting relation is identity relation, as well as the initial arity rewriting relations, and for each $v \in \widetilde{F}_\Sigma(X,\Xi)^{(2)})$ $f(v) = v^{(1)} v^{(2)} f(v_L)$ and for each $\mu \in \widetilde{F}_\Sigma(X,\Xi)^{(1)})$ $f(\mu) = \mu^{(1)} f(\mu_L)$.

## *Renetting Systems*

Let T and S be arbitrary jungles and P a family of sets of positions. We denote

$$T(P \looparrowleft S : *) = \bigcup(v(\mu_i^{(1)} v_i s; \lambda_j^{(1)} \delta_j s) : t = v(\mu_i; \lambda_j \mid i \in \mathcal{I}_v^{UN}, j \in \mathcal{J}_v^{UN}), p(t,\mu_{iL}) \in P,$$

$$p(t, \lambda_{jL}) \in P, t \in T, s \in S, *, v_i s \in s_{outg}, \delta_j s \in s_{ing}).$$

That is $T(P \looparrowleft S : *)$ is the jungle which is obtained by "replacing" (considering conditions *) all the subnets of each net t in T, having the position set in family P, by each net in S.

For given RNS $\mathcal{R}$, jungle S is $\mathcal{R}$-*rewritten* to jungle T (*rewrite result*), denoted $S \to_\mathcal{R} T$, denoted $T = S\varphi$ (the postfix notation is prerequisite), if the following "rewrite" is fulfilled:

$T = \bigcup(S(\mathfrak{p} \looparrowleft (right(r))g) : left(r)$ matches s in $\mathfrak{p}$ by some net substitution mapping $f_{s\mathfrak{p}}, r \in \varphi, g \in G_{s\mathfrak{p}}, \mathfrak{p} \in p(s), s \in S, \mathcal{C}(\mathcal{R}))$, where $G_{s\mathfrak{p}}$'s are sets of net substitution relations. Mapping $f_{s\mathfrak{p}}$ is called *left side substitution relation* and each g in $G_{s\mathfrak{p}}$ is *right side substitution relation*, c.f. under conditional demands $\mathcal{C}(\mathcal{R})$ "extra variables on right-hand sides" *conditional Rewrite Systems* Ohlebusch E (2002). We say that RNS is S-*instance sensitive* (S-INRNS), if for a rule $\varphi \in$ RNS and for each $s \in S$, $\mathfrak{p} \in p(s)$, $G_{s\mathfrak{p}} \neq f_{s\mathfrak{p}}$. Notice that for substitution relations, $\mathcal{C}(\mathcal{R})$ may contain some orders liable to substituting manoeuvre letters in the rewrite process (*substitution order*), especially if rewrite objects have outside loops with the apexes of left sides of pairs in rules or $\mathcal{R}$ is manoeuvre increasing and instance sensitive. Instructions concerning binding right side substitution relations to specific rules in RNS might also have been included in $\mathcal{C}(\mathcal{R})$. We say that $\mathcal{R}$ *matches a rewrite object*, if the left side of a rule preform matches it.

We call RNS *feedbacking in respect to a net*, if while applying a rule in it for that net, elements in the image sets of each right side substitution relation regarding to the preforms in the involving rule overlap that net; feedbacking for a rule is *total*, if the demands concern all elements in the image sets (always total, if the substitution relations are mappings since the image sets are then singletons). If instead of only overlapping, we claim the enclosement condition for elements in the sets of the right side substitution images, feedbacking RNS is *innerly feedbacking* - which is e.g. the case in not instance sensitive RNS´s. If the net in concern of feedbacking is the applicant for RNS, we speak of *self feedbacking*. The form of innerly self feedbacking RNS in respect to a net, say t, where for each rule preform *r* there is in force equation $tr \curlywedge \text{apex}(\text{right}(r)) = t \curlywedge \text{apex}(\text{left}(r))$ ($\curlywedge$ standing for "exclusion"), we name *environmentally saving* in respect to the rewrite object in concern. If all rules in RNS satisfy the feedbacking demands we speak of *thoroughly feedbacking* RNS. It is worth to remind that INRNS´s are capable to join distinct applicable nets.

It is somewhat of worth to mention that RNS´s, not instance sensitive, can own the same rewriting power than INRNS´s, but then we may be compelled to accept infinite number of manoeuvre altering rules – e.g. in the case we have a manoeuvre letter increasing INRNS, where for left side substitution mapping f and right side substitution relation g, $g(y)_L$ overlaps $f(x)_L$ for some manoeuvre letters x≠y (i.e. rewrite results are expected to contain loops) and there is expected to be an infinite number of rewrite objects for which RNS is to be constructed, or if the cardinality of set $\{f(x)_L : x \in X\}$ is infinite.

TRANSDUCER.

For each $\omega \in \Omega$, $i \in \mathcal{I}_\omega$ and $j \in \mathcal{J}_\omega$, let r be a bijection, *RNS-attaching mapping*, joining a set of RNS´s to each triple $(\omega,i,j)$. Let $\mathbb{A} = (F_\Sigma(X,\Xi), \Omega_\mathbb{A} Y_\mathbb{A} \Xi_\mathbb{A})$ be a $\Omega_\mathbb{A} Y_\mathbb{A} \Xi_\mathbb{A}$-algebra, where for each $\omega \in \Omega$

$$\omega^\mathbb{A}: F_\Sigma(X,\Xi)^\alpha \otimes \widetilde{F}_{out\Sigma}(X,\Xi)^\beta \mapsto F_\Sigma(X,\Xi), \text{ where } \alpha = \text{in-rank}(\omega) \text{ and } \beta = \text{out-rank}(\omega),$$

is such an operation relation that

$$\omega^\mathbb{A}(s_{iL}; \lambda_j \mid i \in \mathcal{I}_\omega, j \in \mathcal{J}_\omega) \subseteq \bigcup(s_{iL} r(\omega,i,j): i \in \mathcal{I}_\omega, j \in \mathcal{J}_\omega).$$

$\mathbb{A}$ is called *a renetting algebra*. For any net $t \in F_\Omega(X,\Xi)$ realization $t^\mathbb{A}$ is called R-*transducer* (R-TD) over *RNS-attached family* $R = \{r(\omega,i,j) : \omega \in \Omega \cap L(t), i \in \mathcal{I}_\omega, j \in \mathcal{J}_\omega\}$ of sets of RNS´s and it is also entitled an *interaction* between those RNS´s. We want to notify that for semantic use samples of possible conditions liable to realizations of upstream subnets in carrier nets of transducers may be used to set extra demands for selecting desired operating RNS´s to influence data flows from targeted in-arities.

NORMAL FORM and CATENATION CLOSURE.

$\mathscr{D}(R)$ is the notation for the set of all derivations in TD $R$. If for jungle S and TD $R$, $SR = S$, S is entitled $R$-*irreducible* or of *normal form under* $R$. For the set of all $R$-irreducible nets we reserve the notation IRR($R$). For each jungle S and TD $R$ we denote the following:

$$SR\hat{} = SR^* \cap IRR(R),$$

where $R^*$, *the catenation closure of* $R$, is the transitive closure of the rules in $R$.

Let $R$ be a TD over family $\mathscr{R}$. We define *normal form TD of* $R$, $TD\hat{}$,

$$R\hat{} = R(\mathscr{R} \leftarrow \mathscr{R}\hat{} : \mathscr{R} \in \mathscr{R}).$$

PROBLEM.

*Problem* $\mathfrak{I}$ is a triple $(S, \mathcal{A}, \mathcal{C})$, where the *subject of the problem* S is a jungle, a set of *mother nets*, $\mathcal{A}$ is a recognizer and *limit demands* $\mathcal{C}$ ($\mathcal{C}(\mathfrak{I})$ precise notation, if necessary) is a sample of prerequisites to be satisfied in recognition processes. TD $\mathscr{V}(\mathfrak{I})$ is a *presolution* of problem $\mathfrak{I}$, if $S\mathscr{V}(\mathfrak{I}) \in \mathfrak{L}_\mathcal{A}$, thus $S\mathscr{V}(\mathfrak{I})$ being called a *solution product*, and if furthermore $\mathscr{V}(\mathfrak{I})$ fulfils the demands in set $\mathcal{C}$, $\mathscr{V}(\mathfrak{I})$ is a *solution* of $\mathfrak{I}$. E.g. solution $\mathscr{V}$ may be a system, by which from certain circumstances S, with some limit demands (e.g. the number of the steps in the process) can be built surrounding $S\mathscr{V}$, which in certain state $\alpha(S\mathscr{V})$ (for morphism $\alpha$ of a recognizer) has a capacity characterized by the type of the elements in the final set of the recognizer.

ABSTRACTION CLOSURE THEOREM.

Let A be the set of the denumerable $\theta$-classes, $\mathsf{R}$ a set of RNS´s and

$$\mathscr{R} = \cup(\ \{\ \mathscr{R}_\mathcal{W}^* : \mathcal{W} \text{ is a PRNS of c }\} : \mathscr{R} \in \mathsf{R},\ c \text{ is the centre of Q, Q} \in A\ )$$

a union of macro TD´s liable to A-classes. Then if $\theta$ is the distinctive abstraction relation, pair (A,$\mathscr{R}$) is *a net class rewriting algebra* the operation sets on classes of abstract relation.

CHARACTERIZING GENERALIZED ABSTRACTION.

Let $T \in \{PRNS, GPRNS, CLRNS, GC_dRNS, GCRNS\}$ and let s and t be nets. Then s and t are abstract sisters of type T, if and only if there exist such intervening RNS $\mho_s$ and $\mho_t$ of type T that

$(\exists A_s \in Par(s(\mho_s^{-1})\hat{}))$ and $(\exists A_t \in Par(t(\mho_t^{-1})\hat{}))$ there is a bijection between $A_s$ and $A_t$.

The basis of the previous results leading finally to ubiquitous closure among solving TD´s constituting relation invariability in confluence in other words class rewriting, can be dressed also somewhat more generally:

Let $R$ be a micro TD over set of RNS´s and $W$ a set of intervening nonconditional RNS´s of type T. Then there is such macro TD of $R$, $R_M$, and such set of reversed T-type RNS´s, $W_o$, that we have commutative condition

$$W \hat{} \ R_M \hat{} \ W_o \hat{} \ = R \hat{},$$

and it manifests a natural transformation between Functors determining parallel rewriting.

Conceptual graphs constitute equivalence classes as the form of elements in a closed quotient systems, meaning that parallel transformation applied to those classes inevitably drops images back into the set of those particular classes, which guarantees automated problem solving.

## 2§         Universal Partitioning

In each phase of parallel rule execution resemblance with memory is important due to construction orders to solve original problem. In problem solving commuting requirement is set to guarantee that the initial resemblance between perceived problem and achieved memory counterparts will be preserved while processing memory in order to give a solution from memory back to the original problem, not something totally different object. The commutative property here is generally comparable to the invariability of equivalence relation in parallel transformations or to generalized congruence as well as to derivations in quotient algebras or obedience to confluence, and actually describes closure properties in systems. Interesting is that these properties are related to symmetry and conservation laws, so very essential in nature. Abstraction relations themselves can be regarded as Functors in the set of Categories, intermediating parallel systems.

In this chapter we introduce notion of new intervening RNS, universal partitioning, encompassing the idea of the widest possible memory hunting: each problem is always depending exclusively on its environment and cannot be understand in any other way; even an arbitrary element can only be noted (as symbol level) but not understood feasibly without the idea of its negation. Without the transaction via said universal RNS´s one cannot even think to search solving RNS´s liable to creation of new outside links to the probed objects.

**Definition 2.1.** We denote ILC(s) the set of the inward linkage connections of net s and OLC(s) stands for the outward linkage connections.

**Definition 2.2.** We call cardinality ORN(s) = $\delta_D(s) \cup |\text{ILC}(s)|$ *outward rank number of net* s, $\delta_D$ standing for the cardinality mapping of the positions of the unoccupied arities.

**Definition 2.3.** RNS is *outward rank number saving*, if
$$\text{ORN}(\text{left}(r)) = \text{ORN}(\text{right}(r)),$$
whenever $r$ is a rule preform of the RNS in question. This kind of RNS-type is preserving the character of the realization relations in transformations and quarantining the resemblance in memory hunting between original problem-net and the counterpart in the memory.

**Definition 2.4.** UNIVERSALLY PARTITIONING RNS. For each jungle (here c) we define a *universally partitioning RNS* (UPRNS) $\mathcal{W}$ of that jungle as a RNS fulfilling conditions (i)-(iii):

(i)   $\mathcal{W}$ is thoroughly totally environmentally saving and outward rank number saving,

(ii)  $\mathcal{C}(\mathcal{W}) \supseteq \{L(c) \cap L(c\mathcal{W}\hat{\ }) = \varnothing\}$,

(iii) L(apex(right($r$))) \ $\Xi$ is a singleton and its element is outside L(c), whenever $r \in \varphi$, $\varphi \in \mathcal{W}$, and $\{(\text{left}(r), \text{right}(r)): r \in \varphi, \varphi \in \mathcal{W}\}$ is an injection.

Evidently each GPRNS is UPRNS. An interesting observation can be made: GPRNS are giving resemblances in the memory to nets to be solved such that the mutual order among redexes are preserved, but UPRNS´s give also opportunities to probe the memory among cases where the concerning orders are changed.

**Definition 2.5.** UNIVERSAL ABSTRACTION RELATION. *Universal abstraction relation of type* T ($\in$ITG (={PRNS,GPRNS,CLRNS,UPRNS} (the symbol reserved for this use)))), UAR(T), is defined as GAR, but type T is allowed to be also of type UPRNS.

**Definition 2.6.** UNIVERSAL MACRO TD. *Universal macro of TD* $R$ *over a subset* $\mathcal{T}$ *of* ITG, UMA($\mathcal{T},R$), is TD

$$R(\mathscr{R} \leftarrow \mathscr{R}_{\mathcal{W}_\mathscr{R}} : \mathcal{W}_\mathscr{R} \text{ is of type in } \mathcal{T}, \mathscr{R} \in \mathcal{\mathscr{R}}),$$

where $\mathcal{\mathscr{R}}$ is the set of RNS´s which $R$ is over.

**Definition 2.7.** PARALLEL RNS´s. We generalize in a natural way parallel RNS-definition (cf. Parallel theorem): for abstraction relation $\theta$ of type T ($\in$ITG) we denoted $\mathscr{R}_1 \infty_\theta \mathscr{R}_2$ for $\theta$-parallel RNS´s $\mathscr{R}_1$ and $\mathscr{R}_2$. If $\theta$ has no requirements in addition to its type, say T, we denote for the sake of clarity $\mathscr{R}_1 \infty_T \mathscr{R}_2$. Clearly $\infty_T$ is an equivalence relation in the set of the jungles.

**Definition 2.8.** PARALLEL TD´s. Let $\mathcal{T}$ be a subset of ITG. We say that TD $R$ over set of RNS´s, say $\mathcal{\mathscr{R}}$, and

$$P = \text{UMA}(\mathcal{T}, R(\mathscr{R} \leftarrow \mathscr{P} : \mathscr{P} \infty_T \mathscr{R}, T \in \mathcal{T}, \mathscr{R} \in \mathcal{\mathscr{R}}))$$

are $\mathcal{T}$-*parallel TD´s with each other*, denoted $P \infty_\mathcal{T} R$. Clearly $\infty_\mathcal{T}$ is an equivalence relation in the set of the TD´s and is named *parallel TD-relation over* $\mathcal{T}$, denoted $\infty_{TD}$, if $\mathcal{T}$ is not determined.

**Definition 2.9.** Let $\mathcal{T} \subseteq$ ITG. We say that transformation relations $\rightarrow_{R_1}$ and $\rightarrow_{R_2}$ are $\mathcal{T}$-*parallel* with each other, if $R_1 \infty_\mathcal{T} R_2$.

**Theorem 2.1.** If intervening RNS´s are of the universally partitioning type, parallel RNS´s can be compiled solely in the correspondence with the counterparts in the macro RNS´s in charge.

PROOF. Let s be a net, $\mathcal{W}$ a UPRNS, t be $\mathcal{W}$-irreducible image in catenation closure of $\mathcal{W}$-transformation relation from s and $\mathscr{R}_t$ an arbitrary RNS. Let $\mathscr{R}_s$ be such RNS that

$\exists\, D \in Par(s\mathscr{R}_s)\ \exists$ bijection between sets $\{ORN(\sigma) : \sigma \in L(t)\backslash \Xi\}$ and $\{|OLC(d)| : d \in D\}$.

Therefore $\mathscr{R}_s \infty_{UPRNS} \mathscr{R}_t$. The rest follows from Altering macro RNS-theorem. $\square$.

Next we introduce an extensive example to demonstrate essential features in theorem 2.1

**Example**. We introduce specifically only RNS-rules for initial and final intervening UPRNS´s $\mathcal{W}$ and $\widetilde{\mathcal{W}}$ in the process, each constructed with a little bit different way to depict a variety of alternatives. Let

$\mathcal{W} = \{\varpi_{11}, \varpi_{12} : \text{application order is } \varpi_{11}, \varpi_{12}\}$,

where arity alphabets in tied terms are designated exclusively for the corresponding net representations and

$\varpi_{11} = b(\xi_1 x_1, \xi_i\,;\, \bar{\xi}_j y_j \,|\, i = 2,3, j = 1,2) \rightarrow \beta(\xi_i\,;\, \bar{\xi}_2, \bar{\xi}_j y_j \,|\, i = 1,2, j = 1,3)$,

where for the left side substitution f and for the right side substitution g

$f(x_1) = \bar{\xi}_2 b(\xi_i\,;\, \bar{\xi}_2, \bar{\xi}_1 f(y_1) \,|\, i = 1,2,3)$, notice if in the left side of a rule preform there is a loop structure it can alternatively as here is the case be described as an environmental binding to itself by a substitution,

$f(y_1) = \xi_1 a(\xi_i\,;\, \bar{\xi}_1 \,|\, i = 1,2,3)$

$g(y_1) = \xi_2 a(\xi_1, \xi_2, \xi_3 s\,;\, \bar{\xi}_1)$, $s = \bar{\xi}_3 \beta(\xi_i\,;\, \bar{\xi}_1 g(y_1), \bar{\xi}_2, \bar{\xi}_3 \,|\, i = 1,2)$

$g(y_3) = \xi_3 a(\xi_1, \xi_2 t, \xi_3\,;\, \bar{\xi}_1)$, $t = \bar{\xi}_1 \beta(\xi_i\,;\, \bar{\xi}_1 g(y_1), \bar{\xi}_2, \bar{\xi}_3 g(y_3) \,|\, i = 1,2)$

$\varpi_{12} = a(\xi_1, \xi_i x_i\,;\, \bar{\xi}_1 \,|\, i = 2,3) \rightarrow \alpha(\xi_i x_i\,;\, \bar{\xi}_1 \,|\, i = 1,2,3)$,

where for the left side substitution f and for the right side substitution g

$f(x_2) = \bar{\xi}_1 \beta(\xi_i; \bar{\xi}_j, \bar{\xi}_3 u \mid i = 1,2, j = 1,2)$, $u = \xi_3 a(\xi_i, \xi_2 f(x_2); \bar{\xi}_1 \mid i = 1,3)$

$f(x_3) = \bar{\xi}_3 \beta(\xi_i; \bar{\xi}_1 v, \bar{\xi}_2, \bar{\xi}_3 \mid i = 1,2)$, $v = \xi_2 a(\xi_i, \xi_3 f(x_3); \bar{\xi}_1 \mid i = 1,2)$

$g(x_1) = \bar{\xi}_2 \beta(\xi_i; \bar{\xi}_j, \bar{\xi}_1 p \mid i = 1,2, j = 2,3)$, $p = \xi_2 \alpha(\xi_i g(x_i); \bar{\xi}_1 \mid i = 1,2,3)$

$g(x_2) = \bar{\xi}_1 \beta(\xi_i; \bar{\xi}_j, \bar{\xi}_2 q \mid i = 1,2, j = 1,3)$, $q = \xi_1 \alpha(\xi_1, \xi_i g(x_i); \bar{\xi}_1 \mid i = 2,3)$

$g(x_3) = \bar{\xi}_1 \alpha(\xi_3, \xi_i g(x_i); \bar{\xi}_1 \mid i = 1,2)$

$\tilde{\mathcal{W}} = \{ \varpi_{31}, \varpi_{32} : \text{application order is } \varpi_{31}, \varpi_{32} \}$,

where

$\varpi_{31} = d(\xi_i x_i, \xi_3; \bar{\xi}_1, \bar{\xi}_j y_j \mid i = 1,2, j = 2,3) \to \delta(\xi_i x_i; \bar{\xi}_j y_j, \bar{\xi}_2, \bar{\xi}_4 \mid i = 1,2, j = 1,3)$,

Notice that $\varpi_{31}$ cannot occupy in-arity $\xi_1$ of p for which a gluing point to c is already determined.

$f(x_1) = \bar{\xi}_2 w$

$w = c(\xi_1 \bar{\xi}_2 s, \xi_2 \bar{\xi}_3 s; \bar{\xi}_1, \bar{\xi}_2 \xi_1 s, \bar{\xi}_3 \xi_1 p(\xi_i; \bar{\xi}_1 \mid i = 1,2))$

$s = d(\xi_1 f(x_1), \xi_2 \bar{\xi}_1 s, \xi_3; \bar{\xi}_1 \xi_2 s, \bar{\xi}_j f(y_j) \mid j = 2,3)$

$f(x_2) = \xi_1 s$

$f(y_1) = \xi_2 s$

$f(y_2) = \xi_1 w$

$f(y_3) = \xi_2 w$

$g(x_1) = \xi_1 \bar{\xi}_1 u$

$u = p(\xi_1 \bar{\xi}_3 v, \xi_2 \bar{\xi}_1 t; \bar{\xi}_1 \xi_1 t)$

$t = \delta(g(x_1), \xi_2 g(x_2); \bar{\xi}_1 g(y_1), \bar{\xi}_2, \bar{\xi}_3 g(y_3), \bar{\xi}_4)$

$v = c(\xi_1 \bar{\xi}_3 t, \xi_2; \bar{\xi}_1 \xi_2 t, \bar{\xi}_2, \bar{\xi}_3 \xi_1 u)$

$g(x_2) = \bar{\xi}_1 v$

$g(y_1) = \xi_2 u$

$g(y_3) = \xi_1 v$

$\varpi_{32} = c(\xi_1 x_1, \xi_2 ; \bar{\xi}_j y_j, \bar{\xi}_2 \mid j = 1,3) \to \chi(\xi_i x_i ; \bar{\xi}_j y_j, \bar{\xi}_3 \mid i = 1,2, j = 1,2)$,

$f(x_1) = \bar{\xi}_3 t$

$f(y_1) = \bar{\xi}_2 t$

$f(y_3) = \xi_1 u$

$g(x_1) = \bar{\xi}_2 m$

$g(x_2) = \bar{\xi}_3 m$

$g(y_1) = \xi_1 k$

$g(y_2) = \xi_2 m$

$k = p(\xi_1 \bar{\xi}_1 h, \xi_2 \bar{\xi}_1 m ; \bar{\xi}_1 \xi_1 m)$

$h = \chi(\xi_1 g(x_1), \xi_2 g(x_2) ; \bar{\xi}_1 g(y_1), \bar{\xi}_2 g(y_2), \bar{\xi}_3)$

$m = \delta(\xi_1 \bar{\xi}_1 k, \xi_2 \bar{\xi}_2 h ; \bar{\xi}_1 \xi_2 k, \bar{\xi}_2 \xi_1 h, \bar{\xi}_3 \xi_2 h)$

Starting net q for the process is

$b(\xi_1 \bar{\xi}_2 b(\xi_i ; \bar{\xi}_1 \xi_1 a(\xi_i ; \bar{\xi}_1 \mid i = 1,2,3), \bar{\xi}_2 \mid i = 1,2,3), \xi_i ; \bar{\xi}_1 \xi_1 a(\xi_i ; \bar{\xi}_1 \mid i = 1,2,3) \mid i = 2,3)$

for which we achieve $q \mathcal{W}\hat{\ } R_{\mathcal{W}}\hat{\ } = q R\hat{\ } \widetilde{\mathcal{W}}\hat{\ }$.

**Remark 2.1.** We get the same results for type UPRNS as for earlier represented intervening types in Abstraction closure-theorem and generalized Altering macro RNS-theorem.

**Definition 2.10.** PARTIALLY QUOTIENT ALGEBRA. Let $\mathcal{A} = (A,F)$ be an algebra and $\theta \in Eq(A)$. We say that pair (B,G) is a $\theta$-*partially quotient algebra of* $\mathcal{A}$, if $B = A\theta$ and there is such a bijection $\alpha: F \mapsto G$ that for each f in F there is valid commutation $\theta\alpha(f) \subseteq f\theta$.

**Theorem 2.2.** Let A be a distinctive jungle, G be a set of TD´s, and $\theta$ a distinctive universal abstraction relation of type in ITG over A. Hence net class rewriting algebra is a partially quotient algebra.

PROOF. Let (A,F) be a renetting algebra, and $\alpha : R \mapsto D_R$, $D_R \subseteq \{ P : P\infty_\theta R, P \text{ is a TD} \}$, $R \in F$, be a mapping. Altering macro Clause for type GPRNS and CLRNS Tirri SI (2013) and Remark 2.1 yield $R\theta = \theta P$, whenever $R \in F$ and $P\infty_\theta R$. From the distinctive nature of equivalence classes follows $\alpha$ is a bijection thereby Abstraction closure Tirri SI (2013) yields $(A\theta, \alpha(F))$ is a $\theta$-partially quotient algebra of (A,F). □

**Definition 2.11.** ISOMORPHISM. If net homomorphism is a bijection changing at the maximum symbols of letters without altering ranks of ranked letters or cardinalities of letters or positions in the nets in its domain, we speak of *net isomorphism*.

**Proposition 2.1.** UNIQUENESS OF UAR-CENTER. Centers in the same UAR-class are unique up to net isomorphism.

PROOF. Let us assume in the contradiction that in the same class there are two centers d and d´ not net isomorphic with each other. Then we have two cases: A.) there is a difference in the ranks of d and d´; this however via partition yields inequality in the cardinalities of the classes liable to d and d´, and B.) a difference in cardinalities of positions in d and d´ leads likewise contradiction with our presupposition of the common class. □

**Definition 2.12.** PRECATEGORY. *Category* is a pair of a set of *objects* and a set of relations on that object set.

**Definition 2.13.** FUNCTOR. *Functor* $\Gamma$ is such a relation between precategories: $(A,F) \mapsto (B,G)$, that the following commutation is valid:

   $af\,\Gamma\ = a\Gamma f\,\Gamma$, whenever $a \in A$ and $f \in F$.

We call relation $\Gamma$: object $\mapsto$ object, $\Gamma_{ob}$, the *object projection relation of $\Gamma$*, and relation

$\Gamma$: relation $\mapsto$ relation, $\Gamma_{re}$, the *relation projection relation of $\Gamma$*. It holds a worth to note that Functors are homomorphisms, if the domains and the image sets are algebras. *Normal form $\mathcal{T}$-TD-Functor* $\Gamma$ is such a Functor that the objects are jungles and relations in the pairs of $\Gamma_{re}$ are within each other $\mathcal{T}$-parallel normal form TD transformation relations.

**Proposition 2.2.** Universal abstraction relation of type T ($\in$ITG) is the object projection relation of a normal form T-TD transformation Functor for upside down tree TD´s.

3§     Overlapping Partition Rewriting

1°     *Net Block Homomorphism Deriving Solutions*

We use net block homomorphism as intervening RNS to implement abstract sisters and commutation. First we extend our net presentation by using overlapping cover blocks.

**Definition 3.1.1.** NET NUO-PRESENTATION, a linkage presentation.

For net $t = s(\mu_i; \lambda_j \mid i \in \mathcal{I}_s^{UN}, j \in \mathcal{J}_s^{UN}, C)$ set $\{s, \mu_{iL}, \lambda_{jL} : i \in \mathcal{I}_s^{UN}, j \in \mathcal{J}_s^{UN}\}$ is entitled *block* of t.

Let then $T = \{s, \mu_i, \lambda_j : i \in \mathcal{I}, j \in \mathcal{J}\}$, where $\mathcal{I} \subseteq \mathcal{I}_s$, $\mathcal{J} \subseteq \mathcal{J}_s$ and the indexed nets in T are supposed to occupy indicated arity letters of t in s, we say that $s(\mu_i; \lambda_j \mid i \in \mathcal{I}, j \in \mathcal{J}, C)$, is a *net NUO-representation* of t and we denote $t = s(\mu_i; \lambda_j \mid i \in \mathcal{I}, j \in \mathcal{J}, C)$, and set$\{s, \mu_{iL}, \lambda_{jL} : i \in \mathcal{I}, j \in \mathcal{J}\}$ is entitled its *block*. The block of a letter is the letter itself. The NUO-representation of net t with block D is indicated by D-NUO(t) and NUO(t) is asserted on the establishment for the set of all NUO-representations of t. Notation block(t) is asserted to stand for the family of all block-collections in NUO(t).

Notice that each $\mu_{iL}$ and $\lambda_{jL}$ may be nets in enc(t) not necessarily totally isolated from s, although $\mathcal{I} \cap \mathcal{I}_s^{UN}$ and $\mathcal{J} \cap \mathcal{J}_s^{UN}$ may be nonempty, in other words $\mu_{iL}$ or $\lambda_{jL}$ may overlap net s thus comprising the key feature of NUO-representation. Observe also that $\{s, \mu_i, \lambda_j : i \in \mathcal{I}, j \in \mathcal{J}\}$ is a cover of t, and conversely for any cover of t there is such a NUO-representation that each element in the said cover stands for a net in the block of a net in NUO(t). If set T ($\in$block(t)) is not distinctive, $\cup$(block(p) : p$\in$T) is a genuine subset of block(t). For NUO(t)-representations classes of equal nets [s], s$\in$NUO(t), are defined analogous with t-class definition. Furthermore we obtain block([t]) = $\cup$(block(p) : p$\in$[t]). If net p is in NUO(enc([q])), then q = [q(p | )] = [p(q | )]. Because NUO-representation is covering the net definition, in the following our presumption (if not stated other) is simply to use NUO-representation for nets and assume indexes in nets be subsets of indexes in block elements as described above. For each net t we denote reverse NUO, $NUO^{-1}$ mapping, by asserting: $t = NUO^{-1}(NUO(t))$.

**Definition 3.1.2.** NET BLOCK HOMOMORPHISM.

Let $t = s(\mu_i; \lambda_j \mid i \in \mathcal{I}, j \in \mathcal{J}, C)$, where $\mathcal{I} \subseteq \mathcal{I}_s$, $\mathcal{J} \subseteq \mathcal{J}_s$, be a net in $F_\Sigma(X, \Xi_\Sigma)$ and $D \subseteq F_\Sigma(X, \Xi_\Sigma)$. We define *net D-block homomorphism relation* (D-NBH) h as net homomorphism earlier, but ranked letter rewriting relation $h_\Sigma$ is replaced by *net block rewriting relation* $h_D : D \mapsto h_\Sigma(\Sigma)$ and

$$h(t) = h_D(s)(h(\mu_i); h(\lambda_j) \mid i \in \mathcal{E}_{inh_D(s)}^{UN}, j \in \mathcal{E}_{outh_D(s)}^{UN}).$$

Jungle D is indicated via corresponding NBH h by notation block(h). Relation h is *alphabetically unexpanding* (AlpUnexNBH), if $h(X \cup \Xi_\Sigma) \subseteq Y \cup \Xi_\Omega$, and for each $d \in D$ |rank($h_D(d)$)| is not greater than |rank(d)|. Furthermore we say that h is entitled *right hand side alphabetical* (AlpNBH), if $h(X \cup \Xi_\Sigma) \subseteq Y \cup \Xi_\Omega$, and for each $\sigma \in \Sigma$ $h_\Sigma(\sigma) = \omega(\varepsilon_i ; \varepsilon_j \mid i \in \mathcal{E}_{in}, j \in \mathcal{E}_{out})$, where $\omega \in \Omega$ and $\{\varepsilon_i ; \varepsilon_j \mid i \in \mathcal{E}_{in}, j \in \mathcal{E}_{out}\}$ is an arity alphabet. Net block homomorphism is (*overlapping*) *environment saving*, denoted (D)-(O)ESNBH, and we say that it is *abstracting* (ANBH), if it does not delete the contexts (e.g s in our example) of (overlapping) subnets (e.g. $\in\{\mu_i, \lambda_j : i \in \mathcal{I}, j \in \mathcal{J}\}$) and preserves at least one linkage between the preimage contexts and each of their (overlapping) subnets. Right hand side alphabetical and environment saving net block homomorphism is called *alphabetically abstracting*, abbreviated (D)-AlpANBH. Furthermore we say that NBH is (*overlapping*) *linkage cardinality saving* ((D)-(O)LSNBH), if it is (overlapping) environmental saving and additionally preserves the cardinality of linkages between the preimage contexts and each of their (overlapping) subnets. NBH which is both ANBH and LSNBH is entitled *straightforwardly abstracting* denoted SANBH. Notice that for each net in the preimage domain of D-NBH there is a t-saturating subset of D. In the following we denote (T)NBH for meaning the set of all NBH-relations of type T (defined above).

**Proposition. 3.1.** INVERSE NBH, RESTORING GROUND LEVEL (PERCEPTION)

For each ANBH there is an inverse ANBH.

PROOF. Let h∈D-ANBH. Because h is environment saving its net block rewriting relation $h_D$ is reversible and we can choose such an ANBH, say f, with block(f) = block(Dh) and $h_D^{-1}$ as its net block rewriting relation that hf is an identity mapping in the preimage domain of h. □

**Result. 3.1.1.** Let t be a net and $\mathcal{H}\in Cov(t)$. For each such $Q\subseteq\mathcal{H}$ that $\cap Q \neq \varnothing$, ESNBH h creates $(|Q|-1)\,|rank(\cap Q)|$ new outward links between nets h(a), a∈Q, compared to the counterparts in Q.

**Definition 3.1.3.** ABSTRACTION RELATION VIA NBH.

We extend our notion of abstraction relation to comprise also NBH as an intervening operation type and say that two jungles are in *NBH-abstraction relation* with each other (forming an abstract sister pair), if they are NBH-images of NUO-presentations of the same jungle.

Next theorem will set the general base result that perception regarded as ground basis abstraction of a mother net of given problem can be solved by memory basis abstraction of the same net.

However the question is can we find an implicit solution of an equation, in a resemblance to altering macro RNS theorem, for another of the two abstract sisters (B) when we already have done partition in its origin K in preparation to convert (memory) solution (for the other abstract sister A) to as a solution for K, i.e. to find a macro from found micro of memory solution; micro solutions are easier to find.

**Theorem 3.1.1.** Let $\theta$ be AlpANBH-abstraction relation. For each RNS $\mathscr{R}$ there is RNS $\mathscr{P}$ satisfying commuting equation $\theta\mathscr{R} = \mathscr{P}\theta$.

PROOF. Let (A,B) be an AlpANBH-abstract pair of nets, $\mathcal{W}_1$ and $\mathcal{W}_2$ being of AlpANBH-type intervening NBH´s in concern, A $\mathcal{W}_1$-image of net s and B $\mathcal{W}_2$-image of net t while s and t being NUO-presentations of the same net, say K, with block of s $D_s \subseteq \text{block}(\mathcal{W}_1)$ and block of t $D_t \subseteq \text{block}(\mathcal{W}_2)$. Without a loss of generality we make an assertion:

$s = p(\mu_i; \lambda_j \mid i \in \mathcal{I}_p, j \in \mathcal{J}_p, C)$ and

$t = q(\mu_i; \lambda_j \mid i \in \mathcal{I}_q, j \in \mathcal{J}_q, (\exists\, k \in \mathcal{I}_q^{oc} \cup \mathcal{J}_q^{oc})\, p \in \{\mu_k, \lambda_k\}, C)$.

Let *r* be in A a redex possessing rule preform as a part of a known memory solution RNS for A. Next we construct the following rule preforms:

- micro(*r*):

left side: $\text{apex}(\text{left}(\text{micro}(r)))\mathcal{W}_1 = \text{apex}(\text{left}(r))$ and micro(*r*) matches K

right side : First we define notion net induced AlpANBH: Let t be a net. N-AlpANBH is entitled *t-induced*, denoted N-AlpANBH$^I$(t), if N = $\alpha(\text{rank}(t))$ is a cover of a net homomorphism image of t, where $\alpha$ is a bijection from the rank alphabet to a set of jungles. Because for each [NUO(t)]-class representative AlpANBH-images are equal for the same AlpANBH, we can now choose right(micro(*r*)) to be a net in the preimage domain of an apex(right(*r*))-induced AlpANBH. From the same reasons for the case "*r* is an instance sensitive INRNS-rule", we can choose for each manoeuvre letter x in the domain of right side substitution g of *r* the x-image of right side substitution of micro(*r*) to be a net in the preimage domain of an AlpANBH$^I$(g(x)).

Therefore there is such an AlpANBH, say $\mathcal{W}_{o1}$, that $s\mathcal{W}_1 r = K(\text{micro}(r))\mathcal{W}_{o1}$.

- macro(micro(*r*)) (={ $r_1, r_2$ }) :

First we construct such a new AlpANBH, say $\mathcal{W}_3 : K \mapsto B$ that block($\mathcal{W}_3$) is a partition induced by union block($\mathcal{W}_1$)$\cup$block($\mathcal{W}_2$):

$\{\cap D' \cup \{\cap D'': D' \subset D'', D'' \in P(D)\} : D' \in P(D)\}$, where $D = block(\mathcal{W}_1) \cup block(\mathcal{W}_2)$.

1. first executed rule preform $r_1 : t\mathcal{W}_2 \mapsto (NUO^{-1}(t))\mathcal{W}_3$, the choice guaranteeing in rewriting processes perseverance of analogous environments in $K\mathcal{W}_2$ and K.

2. secondly executed rule preform $r_2$ :

left side: $apex(left(r_2)) = (NUO^{-1}(apex(left(micro(r)))))\mathcal{W}_3$ and $r_2$ matches $(NUO^{-1}(t))\mathcal{W}_3$

right side: Following analogously constructing right side of micro(r) we are free to choose an intervening AlpANBH, say $\mathcal{W}_{o2}$, with the block equal with the block of $\mathcal{W}_{o1}$, and accordingly we choose: $apex(right(micro(r)))\mathcal{W}_{o2} = apex(right(r_2))$.

Therefore $t\mathcal{W}_2 r_1 r_2 = K(micro(r))\mathcal{W}_{o2}$. □

**Result. 3.1.2.** ABSTRACTION RELATIONS INDUCED EQUIVALENCE RELATION OVER THE WHOLE SET OF THE NETS.

Let h be a NBH and T be its preimage domain. h-Abstraction relation $\theta_h$ is in Eq(T), because for each nets s and t in T $(s,t) \in \theta_h$, iff $\exists C_s \in Sat(s)$ and $C_t \in Sat(t)$ that $C_s \cup C_t \subseteq P(T)$. Now we can construct an equivalence relation over the whole set of the nets via AlpUnexNBH-abstraction relations:

$$\cup(\theta_h : h \text{ is a D-AlpUnexNBH}, D \in Par(F_\Sigma(X, \Xi_\Sigma))) \in Eq(F_\Sigma(X, \Xi_\Sigma)).$$

2° *Generating Net Rewriting*

Net block homomorphism execution in the set of nets is more general and powerful than single renetting rule. But as this chapter shows normal form in renetting are able to produce the equal results.

NBH execution can be presented by normal form of rewriting, if at first representation change is effected within domains of net classes so that the inward links of contexts will become outward links to context overlapping subnets as shown inductively in the following definition:

**Definition 3.2.1.** NON-OVERLAPPING NUO-REPRESENTATION RISING MAPPING.

Let $t = s(\mu_i; \lambda_j \mid i \in \mathcal{I}, j \in \mathcal{J}, C)$ be an arbitrary NUO-representation. We define *the non-overlapping NUO-representation rising mapping* in the set of nets, denoted NORNUO, f :

$t \to f(t) = f(s)(\mu_i^{(1)}\mu_i^{(2)}f(\mu_{iL}); \lambda_j^{(1)}\lambda_j^{(2)}f(\lambda_{jL}) \mid i \in \mathcal{I}_{f(s)}^{UN}, j \in \mathcal{J}_{f(s)}^{UN}, C)$ $(\in [t])$, $f(s) = s \mathcal{l} \uplus (\mu_{iL}, \lambda_{jL} : i \in \mathcal{I}, j \in \mathcal{J})$,

where $\mathcal{l}$ is a symbol for net omission and *union* $\uplus(.)$ stands for the net saturated by the set $\{.\}$ of its arguments.

**Definition 3.2.2.** NET BLOCK HOMOMORPHISM RNS. We define for each index of each index set in each net, say i, a frontier letter $x_i$. Let $t = s(\mu_i; \lambda_j \mid i \in \mathcal{I}, j \in \mathcal{J}, C)$. For each consecutive steps in contexts in block D by D-NBH h we establish rules via NORNUO f subject to the same contexts: in t $s \mapsto h(s)$, $s \in D$, corresponds rule

$f(s)(\mu_i \leftarrow \mu_i^{(1)}x_i; \lambda_j \leftarrow \lambda_j^{(1)}x_j \mid i \in \mathcal{I}_{f(s)}^{UN}, j \in \mathcal{J}_{f(s)}^{UN}) \to h(s)(\mu_i \leftarrow \mu_i^{(1)}x_i; \lambda_j \leftarrow \lambda_j^{(1)}x_j \mid i \in \mathcal{E}_{inh_D(s)}^{UN}, j \in \mathcal{E}_{outh_D(s)}^{UN})$,

where the left side manoeuvre alphabet is $\{x_k : k \in \mathcal{I}_{f(s)}^{UN} \cup \mathcal{J}_{f(s)}^{UN}\}$ and $\{x_k : k \in \mathcal{E}_{inh_D(s)}^{UN} \cup \mathcal{E}_{outh_D(s)}^{UN}\}$ is for the right side respectively. The corresponding renetting system is entitled (with a prospective condition set) *net block homomorphism RNS*, shortly NBHRNS. Consequently we obtain the following theorem.

**Theorem 3.2.1.** NBH GENERATION BY RNS-normal form.

For each NBH h there is such a RNS $\mathcal{R}$ that $h = \mathcal{R}\hat{\,}$.

PROOF. We choose NBHRNS as an executing RNS-type. □

The set of the rule preforms of rule r in RNS $\mathcal{R}$ is denoted pre(r) and pre($\mathcal{R}$) = {pre(r) : r ∈ $\mathcal{R}$}.

**Definition 3.2.3.** RENETTING NBH.

Let $\mathcal{R}$ be a RNS and let D = apex(left(pre($\mathcal{R}$)))∪Dom(g), where g is the right side substitution of $\mathcal{R}$. We define *renettig NBH* (RNNBH) h as NBH with net block rewriting relation $h_D$:

apex(left(*r*)) ↦ right(*r*), *r* ∈ pre($\mathcal{R}$),

$$x \mapsto g(x), x \in \text{Dom}(g).$$

Notice that if the left side and right side substitutions are equal, we simply can define

$D = \text{apex}(\text{left}(\text{pre}(\mathscr{R})))$ and $h_D : \text{apex}(\text{left}(r)) \mapsto \text{apex}(\text{right}(r))$, $r \in \text{pre}(\mathscr{R})$.

**Theorem 3.2.2.** RNS-normal form GENERATION BY NBH.

For each RNS $\mathscr{R}$ there is such a NBH h that $\mathscr{R}\hat{\ } = h$ .

PROOF. We choose RNNBH as an executing NBH-type. □

**Theorem 3.2.3.** The operational efficiency of the set of all NBH´s and on the other hand of all RNS´s equates.

PROOF. Combination of generation theorems above. □

As the conclusion we can form quotient algebra with elements being AlpUnexNBH-abstraction relation classes and with operations being bunches of parallel TD´s over RNS normal forms or NBH´s. And if you know a solution for one element in an abstraction relation class you know solutions for all representatives in that particular class. Knowing solutions in the measure of equivalence relation class cardinality is the only necessity in order to get all conceivable solutions. Furthermore obtaining TD-solutions over a sample of abstraction classes for one representative per class results solutions for all routes derived through these classes.

# 4§ N:th Order Net Class Rewriting Systems

## 1° TD-SOLUTION ABSTRACTIONING

**Definition 4.1.1.** TD-ABSTRACTION RELATION.

We define *TD-$\Lambda$-abstraction relation* $\theta_{TD\Lambda}$, $\Lambda \in ITG$, in the set of the elements in abstract algebra $\Lambda$ as follows: Let H and K be two $\Lambda$-derived TD-operations and we mark renetting algebra by $\mathcal{N}$. We define H $\theta_{TD\Lambda}$ K, iff $H^{-\mathcal{N}} \theta_{N\Lambda} K^{-\mathcal{N}}$, where $\theta_{N\Lambda}$ is the $\Lambda$-abstraction relation in the set of the nets and $TD^{-\mathcal{N}}$ stands for the carrier net of TD in concern. If $\Lambda$ is not specified we write simply $\theta_{TD}$ and $\theta_N$ respectively.

**Theorem 4.1.1.** Parallel TD-relation classes are in TD-abstraction relation with each other, iff any of the representatives of them are subject to that.

PROOF. The claim follows from the fact that between each pair of carrier nets of parallel TD-relation class representatives there is an intervening linear alphabetic net homomorphism. $\square$

**Corollary 4.1.1.** Parallel TD-relation classes saturate relevant TD-abstraction relation classes.

PROOF. The claim follows from theorem 4.1.1, because both of the concerning relations are equivalence relations. $\square$

## *CARDINALITIES*

Next we generalize our abstraction relation to cover simultaneously saturated net sets via direct products.

**Definition 4.2.1.** MULTIDIMENSIONAL ABSTRACTION RELATION.

Let $\bar{\Theta}_o \subseteq Eq(F_\Sigma(X,\Xi))$. For each $k \in \mathbb{N}$ and $t \in F_\Sigma(X,\Xi))$ we denote k-level saturation set of t

$$S_k(t) = \{\{s \in \theta F_\Sigma(X,\Xi) : \theta \in \bar{\Theta}_{k-1}\} : s \in F_\Sigma(X,\Xi)\} \cap Sat(t),$$

and define *multidimensional abstraction relation*, $\bar{\Theta}_k$, as the direct product of elements in $\bar{\Theta}_{k-1}$ such that for each s and t in $F_\Sigma(X,\Xi)$  $(s,t) \in \bar{\Theta}_k$  iff

$$\exists (P \in S_k(s), Q \in S_k(t)) \quad (\forall p \in P)(\exists q \in Q) (\exists \theta \in \bar{\Theta}_{k-1}) \quad (p,q) \in \theta$$

and revised for q and p respectively.

Notice that (k-1)-level saturation sets (saturating the classes liable to equivalence relations in the concerned level) saturates each net in k-level saturation set and clearly $\bar{\Theta}_k \in Eq(F_\Sigma(X,\Xi))$. We achieve:

$$(\forall t \in Eq(F_\Sigma(X,\Xi))) \ |t\bar{\Theta}_k| = \prod(\prod((|q\theta| : q \in Q, Q \in S_i(t)) : \theta \in \bar{\Theta}_{i-1}) : i \in \mathbb{N}, i < k),$$

where $\prod$ stands for the multiplication symbol over its argument set.

We denote nest($\mathcal{R}$) the nest of TD $\mathcal{R}$ and let $s_{\mathcal{R}}$ stands for a subnet of net s matched by RNS $\mathscr{R}$. Furthermore $H_{st}$ is reserved to act as the cardinality of the set of the common origins for $\theta_N$-related nets s and t. Because for abstract partially quotient algebra

$$(s\theta_N)\mathcal{R} \subseteq \{B \subseteq a\theta_N : a \in F_\Sigma(X,\Xi)\}, \text{ whenever } \mathcal{R} \text{ is a TD},$$

we obtain the following claim:

**Claim 4.2.1.** If $\bar{\Theta}_o$ is chosen to be a singleton comprising $\theta_N$, we obtain an upper limit for the cardinality of $\theta_N$-class H related $\infty$-class of parallel TD´s in $\theta_{TD}$-class A at k-level

$$\mu(H,A,k) = \oplus(\, (H_{s\mathcal{R}}t : t\,\theta_N\,s_{\mathcal{R}}) \mid t\bar{\Theta}_k \mid : s_{\mathcal{R}}, t \in H^P, P^{-\mathcal{N}} \preccurlyeq \mathcal{R}^{-\mathcal{N}}, P^{-\mathcal{N}} \in \mathrm{sub}(R^{-\mathcal{N}}), \mathcal{R} \in \mathrm{nest}(R), R \in A\,),$$

where $\preccurlyeq$ stands for "next below" and $\oplus$ for the summation over its argument set.

Even though in the cases intervening RNS`s are of PRNS or CLRNS –types the number of the resulted applicant nets as well as the enclosements in them and consequently the number of the left sides of rule preforms in parallel TD´s may be denumerable, so however as because for each net pair (s,t) $H_{st}$ and even the number of the alternatives for the right sides of each rule preform may be unlimited and undenumerable, there may – subject to the cardinality of final states in applied recognizers – exist problems we are not able to determine if they can be comprehensively solved i.e. they are manifesting problems of inconsistent or undecidable nature, cf. "The Undecidable" Davis M (1965), Rosser JB (1936).

2°    Multiple Level Abstraction Algebra,
      Self-Evolving Problem Solving System

In this chapter we iteratively determine PROBLEM SOLVING EVOLUTION structure.

**Definition 4.2.1.** QUOTIENT RELATION.

Let K be a set and E,G $\in$ Eq(K), G$\subseteq$E. Let F be a set of operations on K. We define for each f$\in$F  f(E/G) = {pG : pG$\subseteq$fE, p$\in$K}.

**Definition 4.2.2.** THE FIRST ORDER ABSTRACTION ALGEBRA.

Let A be a jungle, F a set of TD´s, $\bar{\Theta}_k$ multidimensional abstraction relation in A at k-level (k$\in$$\mathbb{N}_o$) subject to $\bar{\Theta}_0$ being a singleton comprising $\theta_N$($\in$ITG) and $\infty_{\bar{\Theta}_k}$ parallel TD-relation in F subject to $\bar{\Theta}_k$. Pair (A$\bar{\Theta}_k$, F$\infty_{\bar{\Theta}_k}$) is called *first order abstraction algebra*, where for each a in A and f



in F is defined $a\bar{\Theta}_k f\infty_{\bar{\Theta}_k} = \{bp : b \in a\bar{\Theta}_k, p \in f\infty_{\bar{\Theta}_k}\}$. If $\theta_N$ is distinct then the apexes of the rule performs in each RNS in TD's are distinct from each other and consequently for each k $\infty_{\bar{\Theta}_k}$ is distinct. Therefore

$a\bar{\Theta}_k f\infty_{\bar{\Theta}_k} \in \{B \subseteq a\theta_N : a \in A\}$.

Next we agree notation for expanding power set definition for multiple powers: For each set K

$(\forall n \in \mathbb{N})\ P^n(K) = P(P^{n-1}(K))$ and $P^0(K) = K$.

Next we define "multiple order abstraction algebra".

Because $\infty_{\theta_N}$ saturates $\theta_{TD}$ by Corollary 4.1, hence we can set the following definition:

**Definition 4.2.3.** SECOND ORDER RELATIONS IN ABSTRACTION ALGEBRA.

Let F be a set of TD's, f, g $\in$ F and k$\in\mathbb{N}_o$. We define quotient relation in F, *second order abstraction relation*, $\bar{\Theta}_{kTD}/\infty_{\bar{\Theta}_k}$ and find out that $f\infty_{\bar{\Theta}_k}\ \bar{\Theta}_{kTD}/\infty_{\bar{\Theta}_k}\ g\infty_{\bar{\Theta}_k}$, if $f\bar{\Theta}_{kTD} g$.

Next we define for each k$\in\mathbb{N}_o$ *second order parallel relation* in $P(F)$, $\infty_{(\bar{\Theta}_{kTD}/\infty_{\bar{\Theta}_k})}$:

$(\forall S, T \subseteq F)\ S \infty_{(\bar{\Theta}_{kTD}/\infty_{\bar{\Theta}_k})} T$, if

$(f\infty_{\bar{\Theta}_k})S\ \bar{\Theta}_{kTD}/\infty_{\bar{\Theta}_k}\ (g\infty_{\bar{\Theta}_k})T$, whenever $(f\infty_{\bar{\Theta}_k}, g\infty_{\bar{\Theta}_k}) \in \bar{\Theta}_{kTD}/\infty_{\bar{\Theta}_k}$.

Now we define binary relation $\odot$ in F:

$(\forall s, t \in F)\ s\odot t = (s^{-\mathcal{N}} t)^{\mathcal{N}}$.

Setting requisite $\theta_N$ is distinct yields $\infty_{\bar{\Theta}_k}$ is distinct and hence we obtain a demonstration of a closure system:

$(\forall s \in S)\ s(\bar{\Theta}_{kTD}/\infty_{\bar{\Theta}_k}) \odot T\infty_{(\bar{\Theta}_{kTD}/\infty_{\bar{\Theta}_k})} = \{h\infty_{\bar{\Theta}_k}\odot t : h\infty_{\bar{\Theta}_k} \in s\theta_{TD}, t \in T\infty_{(\bar{\Theta}_{kTD}/\infty_{\bar{\Theta}_k})}\}$

$\in \{B \subseteq k\in f(\bar{\Theta}_{kTD}/\infty_{\bar{\Theta}_k}) : f \in F\}$,



and we are able to manifest *second order abstraction algebra* ( $F(\bar{\Theta}_{kTD}/\infty_{\bar{\Theta}_k})$ , $\odot P(F)\infty_{(\bar{\Theta}_{kTD}/\infty_{\bar{\Theta}_k})}$ ).

Next we expand the notion of abstraction algebra to multiple orders and first inductively enumerate abstraction relations in nested order.

Let A be a set. For each $n \in \mathbb{N}_0$ and $B \in P^n(A)$ we denote inductively

$$\cup^0(B) = B, \cup^1(B) = \cup(\cup^0(B)) \text{ and } \cup^n(B) = \cup^{n-1}(B).$$

For each $n \in \mathbb{N}_0$ we define $k_n \in \mathbb{N}_0$ and $\bar{\Theta}_{n,k_nTD}$ is such a relation in $P^{n-1}(F)$ that

$$\bar{\Theta}_{0,k_0TD} = \bar{\Theta}_{k_0TD} \text{ and}$$

$$(\forall H, K \in P^{n-1}(F)) \; H \; \bar{\Theta}_{n,k_nTD} \; K, \text{ if } (\cup^n(H))^{\neg\mathbb{N}} \; \bar{\Theta}_{n-1,k_{n-1}TD} \; (\cup^n(K))^{\neg\mathbb{N}}.$$

Furthermore we agree with the notations:

$$\langle \theta/\infty \rangle^{\langle 0,k_0 \rangle} = \bar{\Theta}_{0,k_0TD},$$

$$\langle \theta/\infty \rangle^{\langle 1,k_1 \rangle} = \bar{\Theta}_{1,k_1TD}/\infty_{\bar{\Theta}_{0,k_0TD}},$$

$$\langle \theta/\infty \rangle^{\langle n,k_n \rangle} = \bar{\Theta}_{n,k_nTD}/\infty_{\langle \theta/\infty \rangle^{\langle n-1,k_{n-1} \rangle}}.$$

Next for each $n \in \mathbb{N}_0$ and $k_n \in \mathbb{N}_0$ we keep assumed $\bar{\Theta}_{n,k_nTD}$ are distinctive.

**Definition 4.2.4.** N-LEVEL SOLVING.

Let first $T_n \in P^n(F)$, $n \in \mathbb{N}_0$ and $S_n = S_{n-1} \cup T_n$, $n \in \mathbb{N}$, $S_0 = T_0$.

$\Re^{\langle 0 \rangle}(S_0, k_0) = T_0 \bar{\Theta}_{0,k_0TD}$

$\Re^{\langle 1 \rangle}(S_1, k_1) = T_0 \langle \theta/\infty \rangle^{\langle 1,k_1 \rangle} \odot T_1 \infty_{\langle \theta/\infty \rangle^{\langle 1,k_1 \rangle}} \qquad \in \{B \subseteq k \in q \langle \theta/\infty \rangle^{\langle 1,k_1 \rangle} : q \in F\}$,

$\Re^{\langle 2 \rangle}(S_2, k_2) = \Re^{\langle 1 \rangle}(S_1, k_1) \langle \theta/\infty \rangle^{\langle 2,k_2 \rangle} \odot T_2 \infty_{\langle \theta/\infty \rangle^{\langle 2,k_2 \rangle}} \qquad \in \{B \subseteq k \in q \langle \theta/\infty \rangle^{\langle 2,k_2 \rangle} : q \in P(F)\}$,

.



.
.
.

$$\mathfrak{R}^{\langle n \rangle}(S_n, k_n) = \mathfrak{R}^{\langle n-1 \rangle}(S_{n-1}, k_{n-1})\langle \theta/\infty \rangle^{\langle n,k_n \rangle} \odot T_n \infty_{\langle \theta/\infty \rangle^{\langle n,k_n \rangle}} \in \{B \subseteq k \in q\langle \theta/\infty \rangle^{\langle n,k_n \rangle} : q \in P^{n-1}(F)\}.$$

Finally we are ready to define *n:th order abstraction algebra*

$$(\mathfrak{R}^{\langle n-1 \rangle}(S, k_{n-1})\langle \theta/\infty \rangle^{\langle n,k_n \rangle}, \odot P^n(F) \infty_{\langle \theta/\infty \rangle^{\langle n,k_n \rangle}}), \text{ where } S = \cup(P^i(F) : i = 0,1,...,n-1),$$

because of the closure property:

$$\mathfrak{R}^{\langle n-1 \rangle}(S, k_{n-1})\langle \theta/\infty \rangle^{\langle n,k_n \rangle} \odot P^n(F) \infty_{\langle \theta/\infty \rangle^{\langle n,k_n \rangle}}$$

$$\subseteq \{B \subseteq k \in q\langle \theta/\infty \rangle^{\langle n,k_n \rangle} : q \in P^{n-1}(F)\}$$

$$\subseteq \mathfrak{R}^{\langle n-1 \rangle}(S, k_{n-1})\langle \theta/\infty \rangle^{\langle n,k_n \rangle}.$$

Families $S_n$, $n \in \mathbb{N}_0$, correspond to the mother nets of n-level problems. $\mathfrak{R}^{\langle n-1 \rangle}(S, k_{n-1})\langle \theta/\infty \rangle^{\langle n,k_n \rangle}$ corresponds to the set of the mother nets of the n-level problems and $\odot P^n(F) \infty_{\langle \theta/\infty \rangle^{\langle n,k_n \rangle}}$ to the set of the solutions at the respective level.

Next if we assume mother nets and known solutions be fixed in each level and extend above n:th order processes further exponentially we´ll get autonomous evolution levels and which are of utter importance considering self-developing unrestricted solving processes:

AUTONOMOUS EVOLUTION OF M:TH LEVEL.

For each $m \in \mathbb{N}_0$ we define $n_m \in \mathbb{N}$ and for each ($n_m \in \mathbb{N}$) $k_{n_m} \in \mathbb{N}$ and furthermore inductively by the mapping of solution sets *m:th level autonomous evolution level*

$$\mathcal{EA}_0 = F\langle \theta/\infty \rangle^{\langle 0,k_o \rangle},$$

$$\mathcal{EA}_m \equiv \mathcal{EA}_m(P_{m0}, P_{m1}, ..., P_{mn_m-1}, \mathcal{EA}_{m-1})$$

$$= \mathfrak{R}^{\langle n_m-1 \rangle}(\cup(P_{mi} \subseteq P^i(\mathcal{EA}_{m-1}) : i = 0,1,...,n_m-1), k_{n_m}-1)\langle \theta/\infty \rangle^{\langle n_m, k_{n_m} \rangle}.$$

## Conclusions

This study presents universal partitioning to widen environmental attachments subject to abstract relations yielding universal macros form parallel TD-solutions. Net NUO-presentations are delivered providing more general coverage enabling net block homomorphism to be used for TD-solution generation. A special attention is given to cardinalities of basic solutions. Second order parallel relation is introduced for distinct solution set bases. Finally multiple level abstraction algebra is taken in account for determining self-evolving solving systems. This is reached by tree different stages offering combinational approach in multiple power solution families and iterative solving thus creating solution basis for evolutional levels.

## Acknowledgements

I own the unparalleled gratitude to my family, my wife and five children for the cordial environment so very essential on creative working.

...